\documentclass{article} 
\usepackage{iclr2027_conference,times}

\usepackage{amsmath,amsfonts,bm}

\def\eqref#1{equation~\ref{#1}}

\def\1{\bm{1}}

\DeclareMathAlphabet{\mathsfit}{\encodingdefault}{\sfdefault}{m}{sl}
\SetMathAlphabet{\mathsfit}{bold}{\encodingdefault}{\sfdefault}{bx}{n}

\usepackage{hyperref}
\usepackage{url}
\usepackage{graphicx}
\usepackage{wrapfig}
\usepackage{array}
\usepackage{booktabs}
\usepackage{multirow}
\usepackage{xcolor}
\usepackage[ruled,vlined,linesnumbered]{algorithm2e}
\SetKwInOut{Require}{Require}
\SetKwInOut{Ensure}{Ensure}
\SetKw{Return}{return}
\usepackage{listings}
\newcommand{\pal}{\textbf{\textcolor{blue}{PAL}}}
\usepackage{enumitem}

\title{Plan First, Act Later: Streaming Reasoning for Executable Memory in Robot Control}

\author{Qinghang Liu$^{2,4}$ \quad
Conglin Wang$^1$ \quad
Zongsheng Li$^1$ \quad
Boyong He$^1$ \\
{\bfseries Han Gao$^2$ \quad
Pengcheng Wang$^{3,\dagger}$ \quad
Tao Sun$^{5,\ast}$ \quad
Lige Liu$^{1,\ast}$ \quad
Chang Liu$^{2,4,\ast}$} \\
$^1$JD Logistics \quad
$^2$Peking University \quad
$^3$University of California, Berkeley \quad \\
$^4$PKU-FirstMoveAI Lab \quad
$^5$Fudan University \\
$^\dagger$Project leader \quad
$^\ast$Corresponding author
}

\iclrfinalcopy 
\begin{document}

\maketitle
\lhead{Preprint.}
\begin{abstract}
Long-horizon robot manipulation requires a robot to remember
task-relevant information beyond its short observation window.
Existing memory-augmented policies typically retrieve relevant
information from history and reason over it at execution time,
placing a heavy reasoning burden on the high-level model. To address
this challenge, we propose \pal{} (\textbf{P}lan first, \textbf{A}ct
\textbf{L}ater), a hierarchical memory framework for vision-language
robot manipulation. It maintains a short-term visual buffer, a
semantic-event memory, and an executable \texttt{subtask\_list}. A
high-level vision-language model (VLM) converts streaming visual
observations into the \texttt{subtask\_list}. At execution time, the
VLM selects the currently relevant subtask from the
\texttt{subtask\_list} and passes it to a low-level
vision-language-action (VLA) policy. To teach the model what to write
to memory and when to update it, we introduce a two-stage data
generation procedure that first expands sparse update supervision and
then uses self-labeling with task-grounded verification to generate
training data that better match the state distribution encountered
during execution. We evaluate our approach on memory-dependent
benchmarks and real-world manipulation tasks. On RoboMME, \pal{}
achieves a \textbf{56.05\%} success rate, outperforming the strongest
baseline, FrameSamp-Modul, at 44.51\%. In real-world experiments
across four manipulation tasks, \pal{} reaches \textbf{60.83\%}
overall success, surpassing MemER (45.00\%) and FrameSamp-Modul
(28.33\%). These results show that proactively writing
future-relevant subtasks into hierarchical memory improves
long-horizon visual reasoning and robot control. Code and videos can be found on our website: \url{https://qinghangliu.github.io/pal-webpage/}
\end{abstract}

\begin{figure}[h]
\vskip -0.1in
    \centering
    \includegraphics[width=\textwidth]{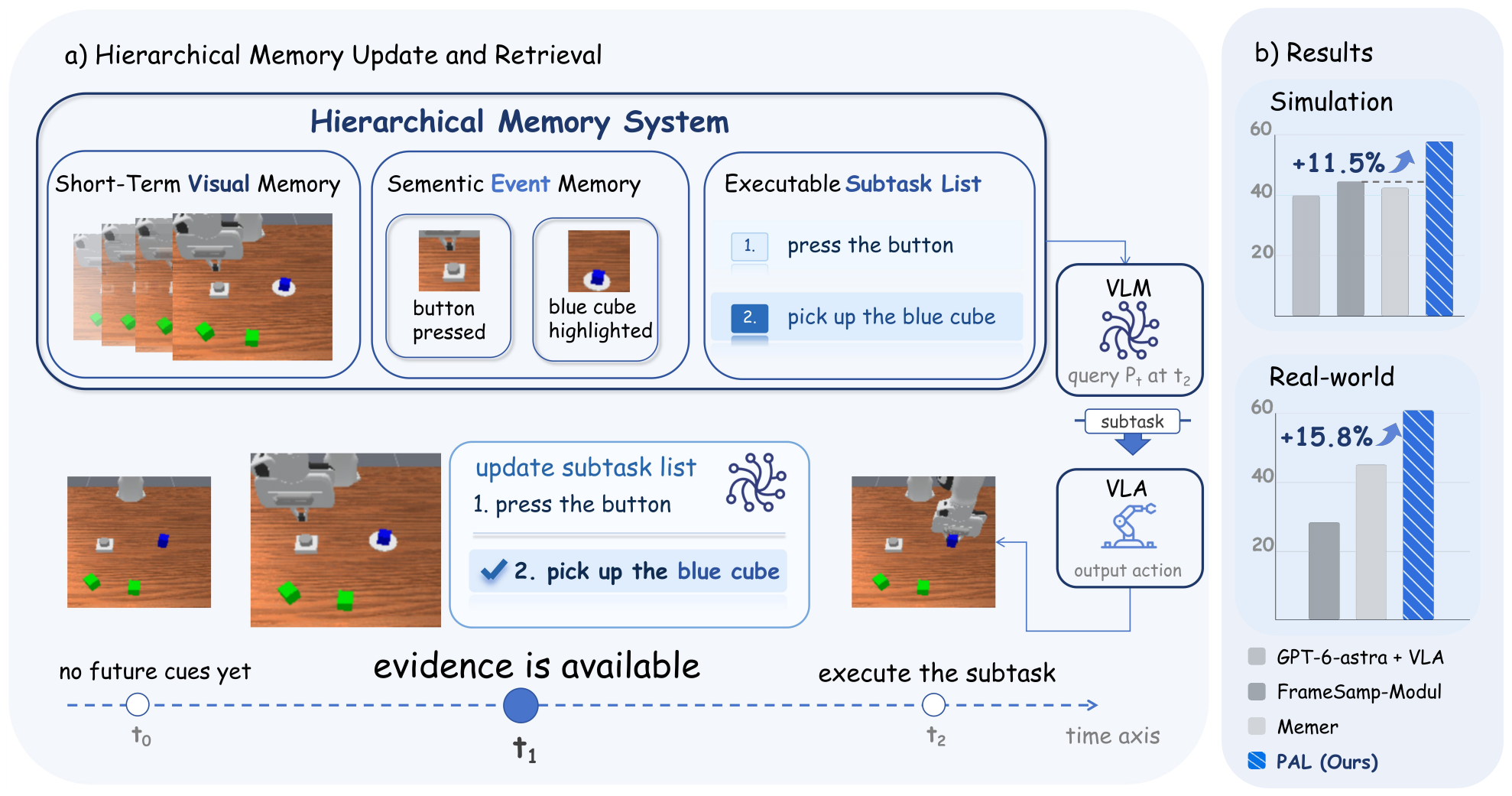}
    \caption{Overview of our pipeline. The high-level VLM incrementally converts
    streaming visual observations into hierarchical memory and retrieves the
    currently relevant subtask for execution by the low-level VLA policy.}
    \label{fig:main-overview}
\vskip -0.1in
\end{figure}

\section{INTRODUCTION}

Vision-language-action (VLA) models have made substantial progress
toward general-purpose robot control
\citep{intelligence2025pi05visionlanguageactionmodelopenworld,
kim2024openvlaopensourcevisionlanguageactionmodel}. By grounding language
instructions in visual observations and actions, VLA provides
a flexible interface for solving diverse manipulation tasks in generalizable open-world. However, many
existing policies still rely on a Markovian or near-Markovian view of the
world: the action to execute is assumed to be recoverable from the current
observation or a short window of recent frames. This assumption becomes fragile
in memory-dependent settings, where successful behavior may depend on historical information observed much earlier. For example, repetitive
manipulation requires remembering progress \citep{yang2026eventvlaeventdrivenvisualevidence,dai2026robommebenchmarkingunderstandingmemory,chen2026rmbenchmemorydependentroboticmanipulation} and active exploration requires
retrieving the location or state of an object discovered earlier when the robot revisits 
the scene \citep{shah2026memoryretrievalvisuomotorpolicies,sridhar2025memerscalingmemoryrobot}.

To address these memory-dependent challenges, recent works augment VLA models
with a memory bank that retains task-relevant information from past
observations. At execution time, the model retrieves relevant entries from the
bank to support stable long-term decision making
\citep{torne2026memmultiscaleembodiedmemory,sridhar2025memerscalingmemoryrobot,
shi2026memoryvlaperceptualcognitivememoryvisionlanguageaction}. Although this
avoids repeatedly processing the raw history, the high-level model must still
search across a large memory bank and infer the next subtask at inference time. Such
banks often contain redundant information, making it difficult for the model to
fetch the required information and lead to more task failures.

One key observation is that: 
\textbf{\textit{Evidence for defining a future goal are available before executing the goal.}}
As illustrated in
Fig.~\ref{fig:main-overview}, the task requires the robot to press the button
and then pick up the highlighted cube. While the robot is still pressing
the button, the model can store the future subtask ``pick up the cube'' in the
subtask list when the cube is highlighted. When the transient highlight
disappears, the robot can simply retrieve the stored subtask for execution,
without re-identifying which cube to pick up. This pattern occurs in many memory-dependent
robotic tasks, where the future plan can be determined once the necessary
evidence becomes available.

Motivated by this view, we propose \textbf{Plan First, Act Later} (\pal{}), a hierarchical
memory pipeline for VLA long-term decision-making. It comprises a three-level memory system:
a short-term visual buffer, a semantic-event memory, and an executable
\texttt{subtask\_list}. These three memory components have distinct roles: the
visual buffer retains recent observations, the semantic-event memory maintains
task-relevant events, and the executable \texttt{subtask\_list} stores
future-relevant subtasks text instructions. At each query, a high-level vision-language model
(VLM) updates the event memory and the \texttt{subtask\_list}, then selects the subtask to execute for the low-level VLA. This design
has two benefits: (1) first, streaming memory updates incrementally compress the visual
history, reducing the context required
for long-horizon reasoning; (2) second, the stored subtasks can be directly retrieved without searching across a large memory bank for subtask inference.

To teach the model what to write to memory and when to update it, we propose 
a two-stage training pipeline. The pipeline combines dense update-timing 
supervision with self-labeling and a task-grounded verifier, allowing the 
training data to better reflect the model's execution rollout distribution.
Evaluation results on RoboMME \citep{dai2026robommebenchmarkingunderstandingmemory}, a benchmark of long-horizon, memory-dependent manipulation tasks, 
show that \pal{} achieves a \textbf{56.05\%} success rate, compared with
44.51\% for the strongest baseline, FrameSamp-Modul. In real-world
manipulation experiments across four tasks, \pal{} reaches \textbf{60.83\%}
overall success, outperforming MemER (45.00\%) and FrameSamp-Modul
(28.33\%), demonstrating the practical benefits of our approach.
Our contributions are as follows:

\begin{itemize}[leftmargin=1em, labelsep=0.3em, topsep=0pt, itemsep=2pt]
\item We introduce \pal{}, a hierarchical memory architecture
    with a streaming-style reasoning mechanism for VLA long-term decision-making. It progressively
    transforms streaming visual observations into semantic events and executable
    subtasks, providing a structured memory for diverse
    long-horizon, memory-dependent tasks.

\item We develop a two-stage training pipeline that enables the model to
    learn reliable hierarchical-memory update behavior. It produces accurate
    and timely memory updates, yielding reliable executable subtask lists for
    long-horizon reasoning.

\item We demonstrate the effectiveness of \pal{} through evaluations on
    long-horizon, memory-dependent manipulation tasks, where it delivers
    consistent gains over competitive VLA and memory-augmented methods in both
    simulated and real-world settings.
\end{itemize}

\section{Related Works}
\paragraph{Vision-language-action models for robot control.}
Recent VLA models have evolved from instruction-conditioned visuomotor policies
into general robot control systems. Early efforts such as OpenVLA
and $\pi_{0.5}$ established a unified interface that maps language instructions
and visual observations to low-level actions
\citep{kim2024openvlaopensourcevisionlanguageactionmodel,
intelligence2025pi05visionlanguageactionmodelopenworld}. More recent systems
improve generalization through larger-scale training, broader robot data, and
continued experience-driven learning
\citep{intelligence2025pi06vlalearnsexperience,
intelligence2026pi07steerablegeneralistrobotic,
xiaomiroboticsteam2026xiaomirobotics1scalingvisionlanguageactionmodels}.
Despite this progress, most VLA policies still operate on the current
observation or a short recent context, motivating hierarchical planning and
memory mechanisms for tasks that require information from earlier history.

\paragraph{Hierarchical planning and subgoal-conditioned control.}
Long-horizon robot tasks are often handled with a hierarchical interface, where
a high-level model selects a subtask or subgoal for a low-level policy to execute
\citep{intelligence2025pi05visionlanguageactionmodelopenworld,shi2025hirobotopenendedinstruction,chen2026rmbenchmemorydependentroboticmanipulation}.
This interface separates selecting the next subtask from generating the
low-level actions that execute it. $\tau_0$-VLA extends this hierarchy with
execution memory and world-model-guided test-time computation, predicting and
comparing candidate subtask outcomes before commitment
\citep{cai2026tau0vlahierarchicalrobotfoundation}. Our work adopts the similar
interface:
the low-level policy executes the current subtask, and the high-level model
maintains the semantic state needed to select future subtasks. However,
the hierarchical interface alone does not specify how to use information from
earlier observations, which motivates memory-augmented VLA methods.

\paragraph{Memory-augmented VLA models and benchmarks.}
Memory-augmented VLA methods address this missing component by storing and
retrieving historical information for high-level decision making. Recent works
use amortized multi-frame context
\citep{jang2025contextvlavisionlanguageactionmodelamortized}, multi-scale
embodied memory \citep{torne2026memmultiscaleembodiedmemory}, experience
retrieval \citep{sridhar2025memerscalingmemoryrobot}, keyframe-based memory
\citep{chen2026nonmarkovianlonghorizonrobotmanipulation}, structured scene
memory \citep{ju2026momagraphstateawareunifiedscene}, and perceptual-cognitive
token memories \citep{shi2026memoryvlaperceptualcognitivememoryvisionlanguageaction}.
HiMe introduces hierarchical embodied memory for long-horizon VLA control,
providing another memory-bank approach
\citep{ji2026himehierarchicalembodiedmemory}.
In parallel, benchmarks such as RoboMME, RMBench, and RoboMemArena highlight
that robotic policies often fail when task success requires retaining and using
historical information
\citep{dai2026robommebenchmarkingunderstandingmemory,
chen2026rmbenchmemorydependentroboticmanipulation,
lei2026robomemarenacomprehensivechallengingrobotic}. Although these methods make
past information accessible, they still require the high-level model to search
a large memory bank at execution time, so the computational cost grows with the
amount of stored history. In contrast, our method allows the model to record a
future-relevant subtask as soon as the supporting evidence is observed, before
the subtask must be executed, and later select it from the editable
\texttt{subtask\_list}. Concurrent work, MaP-WAM
\citep{zhao2026memoryplansworldactionmodeling}, similarly plans ahead
and stores future-relevant information in memory. However, MaP-WAM plans only the
next step, whereas our editable \texttt{subtask\_list} retains all inferred
future subtasks and serves as a persistent component of the memory system.

\section{Preliminaries}
\label{sec:preliminaries}
\paragraph{Problem setup.}
We consider memory-dependent VLA tasks in which the
correct behavior at timestep $t$ may depend on information that is not
visible in the current observation. Let $L$ denote the language instruction,
$o_t$ the visual observation, $a_t$ the low-level action, and
$H_t=\{(o_i,a_i)\}_{i=1}^{t}$ the raw interaction history. Long-horizon tasks
are increasingly challenging for a fully history-based policy
$\pi(a_t\mid H_t,L)$ since the visual context grows with time and contains
many redundant frames.

\paragraph{Hierarchical memory formulation.}
We organize task memory at three temporal scales. The short-term visual buffer
$\mathcal{V}_t$ stores recent RGB images from the current observation window.
The other two components are ordered lists of natural-language entries. The
semantic-event memory $\mathcal{E}_t$ records events that have occurred in the
environment and may be relevant to the task, with one event per entry. The
executable memory $\mathcal{P}_t$ is the \texttt{subtask\_list}, and each of
its entries is one subtask. An example is shown in
Fig.~\ref{fig:main-overview}. Together, these components form the hierarchical memory state
\[
\mathcal{M}_t :=
(\mathcal{V}_t,\mathcal{E}_t,\mathcal{P}_t),
\]

At each query, the VLM uses the incoming visual observations and the hierarchical memory to
update either list and to select the currently subtask $\ell_t$, which
is passed to the low-level policy:
\[
a_t=\pi_{\mathrm{VLA}}(o_t,\ell_t,L).
\]
\paragraph{Plan first, Act later.}
Our approach to constructing the subtask list is inspired by a simple temporal
separation: when visual evidence reveals that a subtask will be needed before
it is time to execute it, the model can store the subtask for later execution.
Formally, a subtask written to memory at $t_1$ can be
selected as $\ell_{t_2}$ for execution at a later time $t_2$:
\[
\mathcal{M}_{t_1+\Delta t}=g_{\theta}
\bigl(\mathcal{M}_{t_1},o_{t_1:t_1+\Delta t},L\bigr),
\qquad
\ell_{t_2}\in\mathcal{P}_{t_1+\Delta t},
\quad t_1\leq t_2.
\]

\section{Methodology}
\label{methodology}

\subsection{Overview}
Our methodology has two parts. Section~\ref{sec:streaming-memory-reasoning}
describes the memory construction and querying mechanism used whenever the VLM
runs, including both training rollouts and inference. Section~\ref{sec:two-stage-training}
then describes how we train the VLM to acquire this capability: learning what
to write to memory, when to update it, and which stored subtask to execute.

\subsection{Stream-based Memory Reasoning}
\label{sec:streaming-memory-reasoning}
\begin{wrapfigure}{r}{0.60\textwidth}
    \vspace{-8pt}
    \centering
    \includegraphics[width=\linewidth]{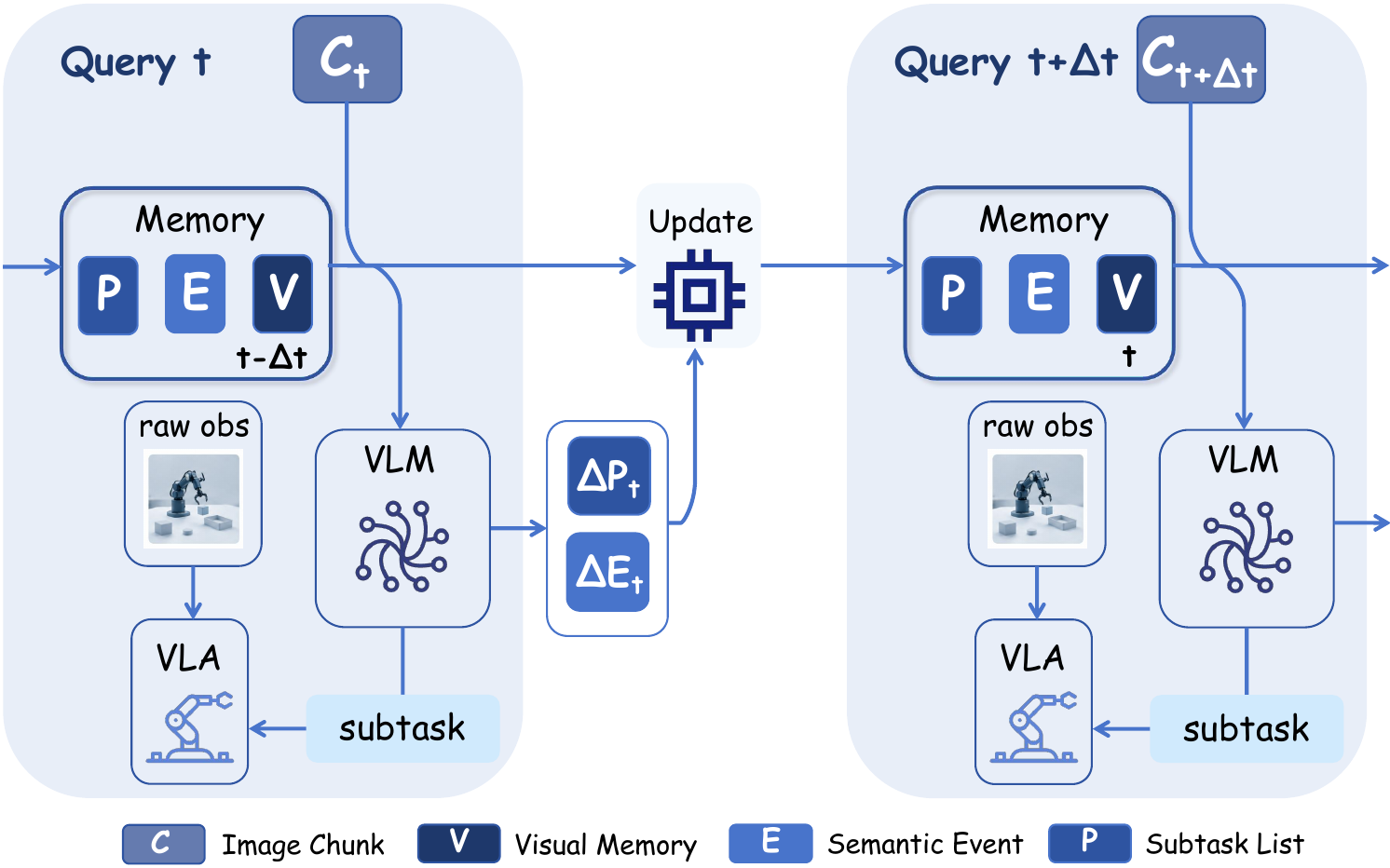}
    \caption{Streaming memory update.}
    \label{fig:memory-update-process}
    \vspace{-8pt}
\end{wrapfigure}

Following the three-level hierarchical memory formulation introduced in
Sec.~\ref{sec:preliminaries}, we propose a streaming-style reasoning pipeline
that processes history as a stream of image chunks, as illustrated in
Fig.~\ref{fig:memory-update-process}. For a query at time $t$, let
$\mathcal{C}_t=o_{t-\Delta t:t}$ denote the observations collected since the
previous query. The VLM takes the previous hierarchical memory
$\mathcal{M}_{t-\Delta t}$, the current observation chunk $\mathcal{C}_t$, and
the task instruction $L$ as input. The visual buffer stored in
$\mathcal{M}_{t-\Delta t}$ is combined with $\mathcal{C}_t$ and downsampled to
a fixed number of frames before being passed to the VLM. The VLM then outputs
three fields. The quantity
$\Delta\mathcal{E}_t$ is a newly predicted event, while $\Delta\mathcal{P}_t$
is a newly predicted subtask to be appended to the subtask list. The third
output, $\ell_t$, is the current subtask passed to the low-level VLA. Because
image chunks arrive incrementally, the evidence available at a query may be
insufficient to determine whether a new event has occurred or which future
subtask should be appended. In this case, the VLM outputs an empty
$\Delta\mathcal{E}_t$ or $\Delta\mathcal{P}_t$, respectively, and the
corresponding memory is retained until further observations become available.
The memory transition is written as
\[
\bigl(\Delta\mathcal{E}_t,\Delta\mathcal{P}_t,\ell_t\bigr)
=g_{\theta}\bigl(\mathcal{M}_{t-\Delta t},\mathcal{C}_t,L\bigr),
\]
For ordered sequences, $\operatorname{Append}(A,B)$ denotes appending $B$ to
the end of $A$. The subtask list is updated as
\[
\mathcal{P}_t=
\operatorname{Append}\left(\mathcal{P}_{t-\Delta t},\Delta\mathcal{P}_t\right).
\]
If $\Delta\mathcal{P}_t$ is non-empty, the previous event memory
$\mathcal{E}_{t-\Delta t}$ is cleared; otherwise, it is retained. Let
$\widetilde{\mathcal{E}}_{t-\Delta t}$ denote this retained or reset event
memory:
\[
\widetilde{\mathcal{E}}_{t-\Delta t}=\begin{cases}
\emptyset, & \text{if } \Delta\mathcal{P}_t\neq\emptyset,\\
\mathcal{E}_{t-\Delta t}, & \text{otherwise}.
\end{cases}
\]
The event memory is updated by appending the predicted event:
\[
\mathcal{E}_t=
\operatorname{Append}\left(\widetilde{\mathcal{E}}_{t-\Delta t},
\Delta\mathcal{E}_t\right).
\]
The visual buffer is cleared after a memory update; otherwise, the new chunk is
appended to the previous visual buffer:
\[
\mathcal{V}_t=
\begin{cases}
\emptyset, & \text{if } \Delta\mathcal{E}_t\neq\emptyset \text{ or }
\Delta\mathcal{P}_t\neq\emptyset,\\
\operatorname{Append}\left(\mathcal{V}_{t-\Delta t},\mathcal{C}_t\right),
& \text{otherwise}.
\end{cases}
\]
Clearing the buffer after a memory update prevents visual evidence from
accumulating excessively and making subsequent reasoning harder. After the
subtask list is updated, the event buffer is also cleared so that subsequent
event recognition starts relative to the updated list.
The complete update procedure for one query is summarized in
Algorithm~\ref{alg:streaming-memory-update} in the appendix.

\subsection{Two-stage training with self-labeling and verification}
\label{sec:two-stage-training}

We propose a two-stage training procedure to teach the model what to write to
the subtask list and when to update it, while jointly training it to predict
the current subtask at each query. Stage 1 augments the training data by
creating additional training examples from each sparse annotation. However,
this augmentation changes the training data
distribution, introducing a mismatch with the model's inference distribution.
We therefore introduce Stage 2 to regularize the model's memory-update behavior
and reduce this mismatch.

\paragraph{Stage 1: Dense update-timing supervision.}
\begin{figure*}[t]
    \centering
    \includegraphics[width=\textwidth]{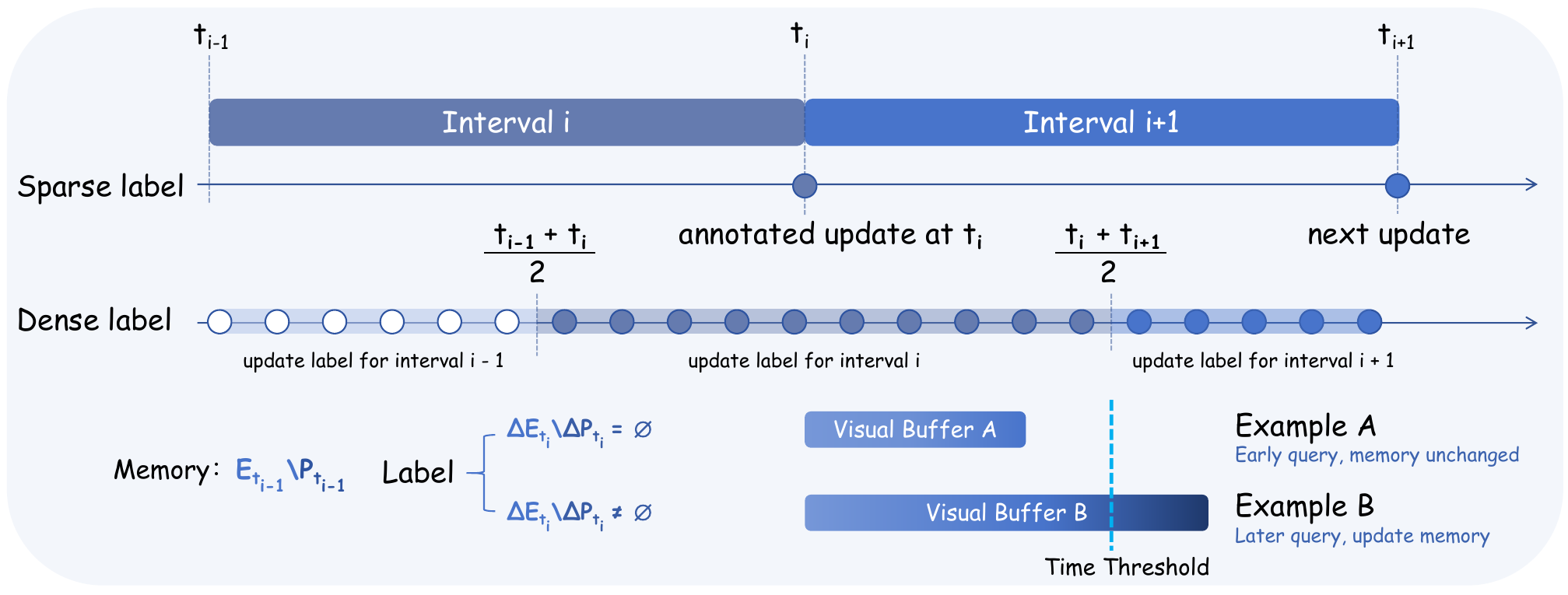}
    \caption{Dense update-timing supervision. A sparse update annotation is
    expanded into multiple training samples with visual buffers of different
    lengths, exposing the model to different levels of evidence around the
    reference update boundary.}
    \label{fig:dense-update-labels}
\end{figure*}

As illustrated in Figure~\ref{fig:dense-update-labels}, we first construct a
dataset of sparse annotations; details of the simulation and real-world dataset
construction are provided in Appendices~\ref{app:dataset-construction} and
\ref{app:real-data-cleaning}, respectively. Each sparse annotation is provided
at a single timestep $t_i$. We denote the training labels for the new event and subtask at this
timestep by $\Delta\mathcal{E}^{\mathrm{ref}}_i$ and
$\Delta\mathcal{P}^{\mathrm{ref}}_i$, respectively. Such sparse
annotation is data-inefficient because it
provides only one supervision point for each update. Moreover, visual evidence
changes only slightly between neighboring queries, making the update and
no-update cases difficult for the model to distinguish.

To address this issue, we sample multiple query timesteps $t$ at regular 
intervals from $[t_{i-1}, \frac{t_i + t_{i+1}}{2})$. For each query, we construct $\mathcal{V}_t$ from the observations accumulated since the preceding update up
to timestep $t$, so later queries contain more visual evidence.
For each sampled query timestep, we assign $\emptyset$ to the event and
subtask-list targets when $t < \frac{t_{i-1}+t_i}{2}$, and assign
$\Delta\mathcal{E}^{\mathrm{ref}}_i$ and
$\Delta\mathcal{P}^{\mathrm{ref}}_i$ when
$t \geq \frac{t_{i-1}+t_i}{2}$. 
We use this simplified assignment as a practical rule: queries presumed to have
insufficient information for an update receive empty targets, whereas later
queries receive the corresponding reference event and subtask labels.
The effectiveness of this simple approximation is further examined in
Sec.~\ref{sec:ablation}.
In this way, the single annotation at $t_i$
is expanded into annotations over a range of query timesteps. The model is fine-tuned on these densified
labels to learn the memory update schema and approximate update timing from the
reference teacher's distribution. 

\paragraph{Stage 2: Self-labeling with task-grounded verification.}
Despite this supervision, the Stage 1 model may still produce updates too frequently or at
imprecise times. In particular, it may update before enough evidence has
accumulated or repeat content that is already in memory. Stage 2 is designed to
address these issues through task-grounded verification.

Concretely, we roll out the Stage 1 model on the training set, maintaining the
memory states generated by the model. Let
$\frac{t_{i-1}+t_i}{2}\leq t<\frac{t_i+t_{i+1}}{2}$ be the timing band of the
$i$-th reference update, whose reference event and subtask-list increments are
$\Delta\mathcal{E}^{\mathrm{ref}}_i$ and $\Delta\mathcal{P}^{\mathrm{ref}}_i$.
Let $\mathcal{X}\in\{\mathcal{E},\mathcal{P}\}$ stand for either memory. For each
prediction, the verifier checks the corresponding reference label and whether that memory
already stores it:
\[
\Delta\mathcal{X}_t\leftarrow\emptyset,
\quad \text{if }\Delta\mathcal{X}_t\neq\Delta\mathcal{X}^{\mathrm{ref}}_i
\text{ or }\Delta\mathcal{X}_t\in\mathcal{X}_{t-\Delta t}.
\]
Thus, an incorrect prediction, as well as one that duplicates an entry already
stored in memory, is replaced with $\emptyset$, while the other prediction is
unaffected. This output is then used to update the memory following the update
mechanism in Section~4.2, and the updated memory state $\mathcal{M}_t$ is the
input of the next query. If
\[
t\geq\frac{t_i+t_{i+1}}{2}
\quad\text{and}\quad
\Delta\mathcal{X}^{\mathrm{ref}}_i\notin\mathcal{X}_t
\text{ for some }
\mathcal{X}\in\{\mathcal{E},\mathcal{P}\},
\]
the rollout is marked as failed, since the model has not produced and stored the
$i$-th reference event or subtask-list update by the deadline. We retain only
rollouts in which the verifier accepts an update at every required update point.
The resulting update annotations remain sparse, and their timing is determined
within a limited range by the VLM. As a result, the data distribution is closer
to the model's output distribution and is easier for the model to learn. 
The model is then fine-tuned on these self-labeled trajectories. The complete
procedure is given in Algorithm~\ref{alg:stage-two-self-labeling} in the
appendix.

\section{Experiments}
\label{sec:experiments}
This section answers the following questions:
\begin{itemize}
    \item By how much does \pal{} improve success rates on memory-dependent manipulation tasks compared with baselines?
    \item Why does \pal{} perform better than baselines on long-horizon tasks?
    \item How do the two training stages contribute to \pal{}'s overall performance?
    \item How does \pal{} perform in real-world robotic manipulation tasks compared with baselines?
\end{itemize}
\subsection{Simulated experiments}

\paragraph{RoboMME benchmark and baselines.}
We evaluate our model on the RoboMME benchmark \citep{dai2026robommebenchmarkingunderstandingmemory}, which is a collection of long-horizon, memory-dependent tasks for robotic manipulation.
We compare \pal{} with five baselines. \textbf{$\pi_{0.5}$} uses only the current
observation and task instruction, providing a memory-free reference without
access to observation history. \textbf{FrameSamp-Modul}~\citep{dai2026robommebenchmarkingunderstandingmemory}
uniformly samples
frames from the observation history and injects their visual features into the
policy through feature modulation, providing a perceptual-memory baseline
without explicit high-level reasoning. \textbf{MemER}~\citep{sridhar2025memerscalingmemoryrobot}
retrieves relevant
experience from the observation history and uses a VLM to produce a grounded
subtask for the low-level VLA. \pal{} adopts the same high-level--low-level control
interface, and the low-level VLA used by MemER is identical to that of \pal{}.
\textbf{EventVLA}
\citep{yang2026eventvlaeventdrivenvisualevidence} is an end-to-end VLA that
learns to retain sparse, task-relevant keyframes in its Keyframe Evidence
Memory. It provides a complementary comparison between \pal{}'s semantic,
executable memory and learned visual evidence memory. Finally,
\textbf{GPT-6-astra + VLA}
replaces \pal{}'s fine-tuned high-level VLM with GPT-6-astra while using the same
low-level VLA, and represents its memory as a language description of the
current environment from which it predicts the next subtask.
Table~\ref{tab:simulation-main-results} reports group-level means. The full
per-task breakdown and the implementation and baseline details are provided in
Appendices~\ref{app:simulation-results} and~\ref{app:simulation-setup},
respectively.

\begin{table}[h]
\centering
\caption{Mean success rates (\%) on RoboMME task groups. \textbf{Short-video}
and \textbf{Long-video} infer an executable subtask list from conditioning
videos of different lengths; \textbf{In-context} adapts it from observations
during the episode; and \textbf{Episode-start} has all information needed to
infer the complete list at the beginning of the episode. The AVG column reports
mean $\pm$ standard deviation when multiple evaluation seeds are available.}
\label{tab:simulation-main-results}
\small
\setlength{\tabcolsep}{3.2pt}
\begin{tabular}{lccccc}
\toprule
\textbf{Method} & \textbf{Short-video} & \textbf{Long-video} &
\textbf{In-context} & \textbf{Episode-start} & \textbf{AVG} \\
\midrule
\multicolumn{6}{l}{\textbf{\textit{Baselines}}} \\
$\pi_{0.5}$ & 10.67 & 28.45 & 11.72 & 36.15 & $17.93 \pm 0.61$ \\
EventVLA & 27.60 & 19.00 & 32.50 & 21.33 & $26.57 \pm 0.13$ \\
FrameSamp-Modul & 41.87 & 46.00 & 27.05 & 72.96 & $44.51 \pm 0.77$ \\
MemER & 37.52 & 28.00 & 41.00 & 65.11 & $42.38 \pm 0.33$ \\
GPT-6-astra + VLA & 31.71 & 66.00 & 28.00 & 56.67 & $39.75$ \\
\midrule
\multicolumn{6}{l}{\textbf{\textit{Streaming Reasoning}}} \\
Stage 1 Only & 31.43 & 17.40 & 32.50 & 70.94 & $37.35 \pm 1.77$ \\
Sparse Label & 36.74 & 30.44 & 38.64 & 73.87 & $43.39 \pm 2.26$ \\
\textbf{\pal{} (ours)} & \textbf{45.26} & \textbf{81.60} & \textbf{46.20} &
\textbf{77.33} & $\mathbf{56.05 \pm 1.38}$ \\
\bottomrule
\end{tabular}
\end{table}
\paragraph{Results and analysis.} The main results are summarized in Table~\ref{tab:simulation-main-results},
with complete per-task results provided in Appendix~\ref{app:simulation-results}.
We analyze these results through the following research questions.

\noindent\textbf{Q1: By how much does \pal{} improve success rates on
memory-dependent manipulation tasks compared with baselines?}

\pal{} achieves an overall success rate of $56.05\%$, improving upon the strongest
baseline, FrameSamp-Modul ($44.51\%$), by $11.54$ percentage points. We focus
the comparison on
MemER and GPT-6-astra + VLA because, like \pal{}, both use a hierarchical control
structure in which a high-level model generates a subtask for a low-level VLA.
\pal{} outperforms MemER ($42.38\%$) and GPT-6-astra + VLA ($39.75\%$) by $13.67$
and $16.30$ percentage points, respectively, and achieves the highest success
rate on 4 of the 16 RoboMME tasks.

\noindent\textbf{Q2: Why does \pal{} perform better than baselines on
long-horizon tasks?}

We attribute \pal{}'s gains to two complementary components: the executable
subtask list decouples subtask inference from execution-time activation, while
streaming memory updates reduce the visual context and VLM capacity required
for long-horizon reasoning.

The \textbf{Episode-start} group directly evaluates the \textit{benefit of constructing}
the subtask list ahead of execution, as all information needed to infer the
complete list is available at the beginning of the episode. \pal{}
achieves $77.33\%$ on this group, outperforming the strongest baseline,
FrameSamp-Modul ($72.96\%$), by $4.37$ percentage points. For a more direct
comparison, MemER and GPT-6-astra + VLA also use a high-level model to generate
a subtask for the low-level VLA. However, \pal{} constructs the complete subtask
list in advance, whereas MemER generates a subtask only when it is needed. \pal{}
outperforms MemER ($65.11\%$) by $12.22$
percentage points, indicating the benefit of planning future subtasks ahead of
execution rather than inferring them on demand.

The remaining groups further demonstrate the \textit{benefit of streaming memory}
updates. \pal{} achieves success rates of $45.26\%$, $81.60\%$,
and $46.20\%$ on these groups, improving over FrameSamp-Modul ($41.87\%$,
$46.00\%$, and $27.05\%$) by $3.39$, $35.60$, and $19.15$ percentage points,
respectively. The largest gain is on \textbf{Long-video}, where the conditioning stream
is longest. \pal{}'s streaming updates compress this visual history into a compact
subtask list, so the VLM reasons over a smaller context rather than the full
video.

\subsection{Ablation Study}
\label{sec:ablation}

To validate the effectiveness of our two-stage training approach, we conduct ablation experiments on the RoboMME benchmark.
We compare four variants: (1) \textbf{Sparse Label}, trained on data that
annotate event or subtask-list updates only at the end of each evidence chunk;
(2) \textbf{Stage 1 Only}, which uses only the densified supervision from the
reference teacher without self-labeling;
(3) \textbf{Update Gate}, which uses the Stage 1 checkpoint together with an
additional ground-truth update gate provided by the environment at inference
time. Similar to the gate in Sec.~4.3, it opens an update slot when each
interval reaches $80\%$ of its duration; the slot remains open until the VLM produces its first non-empty update, and any update outside the active slot is discarded. (4)
\textbf{\pal{}} is our complete approach with both stages and streaming
updates. All of them use the same hierarchical memory architecture and low-level VLA policy as \pal{}. The only
 difference is the training procedure and the update gate used at inference time.

\begin{table*}[t]
\centering
\caption{Subtask-list update behavior. For each task, we report the average
number of \texttt{subtask\_list} updates per episode (Update freq.) and the
accuracy of the final list (Final acc.). Parentheses report the relative change in update frequency from ground truth.}
\label{tab:ablation-update-frequency}
\small
\setlength{\tabcolsep}{2.5pt}
\resizebox{\textwidth}{!}{%
\begin{tabular}{ccccccccc}
\toprule
& \multicolumn{2}{c}{PatternLock} & \multicolumn{2}{c}{RouteStick} &
\multicolumn{2}{c}{VideoPlaceOrder} & \multicolumn{2}{c}{VideoPlaceButton} \\
\cmidrule(lr){2-3}\cmidrule(lr){4-5}\cmidrule(lr){6-7}\cmidrule(lr){8-9}
Variant & Update freq. & Final acc. & Update freq. & Final acc. &
Update freq. & Final acc. & Update freq. & Final acc. \\
\midrule
\multirow{2}{*}{Sparse Label} & 2.68 & \multirow{2}{*}{22.00\%} &
3.34 & \multirow{2}{*}{50.00\%} & 0.00 & \multirow{2}{*}{0.00\%} &
0.00 & \multirow{2}{*}{0.00\%} \\
& (-14.6\%) & & (-8.7\%) & & (-100.0\%) & & (-100.0\%) & \\
\midrule
\multirow{2}{*}{Stage 1 Only} & 4.32 & \multirow{2}{*}{24.00\%} &
5.28 & \multirow{2}{*}{30.00\%} & 1.34 & \multirow{2}{*}{18.00\%} &
1.38 & \multirow{2}{*}{18.00\%} \\
& (+37.6\%) & & (+44.3\%) & & (+34.0\%) & & (+38.0\%) & \\
\midrule
\multirow{2}{*}{Update Gate} & 3.10 & \multirow{2}{*}{78.00\%} &
3.66 & \multirow{2}{*}{82.00\%} & 1.00 & \multirow{2}{*}{80.00\%} &
1.00 & \multirow{2}{*}{78.00\%} \\
& (-1.3\%) & & (0.0\%) & & (+0.0\%) & & (+0.0\%) & \\
\midrule
\multirow{2}{*}{Ground truth} & 3.14 &
\multirow{2}{*}{\rule{1.5em}{0.4pt}} & 3.66 &
\multirow{2}{*}{\rule{1.5em}{0.4pt}} & 1.00 &
\multirow{2}{*}{\rule{1.5em}{0.4pt}} & 1.00 &
\multirow{2}{*}{\rule{1.5em}{0.4pt}} \\
& (0.0\%) & & (0.0\%) & & (0.0\%) & & (0.0\%) & \\
\midrule
\multirow{2}{*}{\pal{}} & 2.94 & \multirow{2}{*}{54.00\%} &
3.46 & \multirow{2}{*}{68.00\%} & 1.04 & \multirow{2}{*}{84.00\%} &
1.02 & \multirow{2}{*}{80.00\%} \\
& (-6.4\%) & & (-5.5\%) & & (+4.0\%) & & (+2.0\%) & \\
\bottomrule
\end{tabular}
}
\end{table*}
\noindent\textbf{Q3: How do the two training stages contribute to \pal{}'s overall performance?}
Table~\ref{tab:simulation-main-results} reports group-level averages, with the
full per-task results in Appendix~\ref{app:simulation-results}. Stage 1 Only
achieves $37.35\%$ overall success, Sparse Label improves to $43.39\%$, and \pal{}
reaches $56.05\%$. To explain these differences, we examine the update behavior
in Table~\ref{tab:ablation-update-frequency} on four representative streaming-reasoning
tasks. PatternLock and RouteStick involve a large number of list updates as
 evidence arrives, whereas VideoPlaceOrder and VideoPlaceButton require evidence to be
accumulated over a long video stream. Comparing Stage 1 Only with Update Gate,
we find that applying the update gate greatly improves final-list accuracy. This is
because Update Gate provides the model with approximate update timings, thereby
decoupling when to update from what to update. The improvements on the final-list accuracy implies that Stage 1 helps the
model learn what content to update, but not when to update it.

We then calculate the update frequency of each variant. As expected,
Stage 1 Only tends to update the executable list too
often, since the model is trained with a dataset that over densifies the update timing.
Sparse Label under-updates and even fails to update any subtask list for Long-video tasks,
which implies sparse supervision is insufficient for learning the correct update behavior.
\pal{}'s update frequency is closer to ground truth and yields higher final-list
accuracy. This suggests that Stage 2 regularizes the overly frequent update
behavior learned in Stage 1, producing more accurate memory states for
long-horizon reasoning.

\subsection{Real-world experiments}

We evaluate our pipeline in the real world using a Franka Research 3 robot arm.
The evaluation comprises four tasks adapted from their simulated counterparts,
each targeting a distinct memory capability. PickXtimes requires the robot to
pick a green box a specified number of times, testing its ability to track task
progress. In VideoUnmask, the model observes a video showing the layout of
colored cubes and must subsequently point to the requested cube. VideoPlaceOrder
tests long-video reasoning by requiring the model to recall where a target box
was placed earlier in the video. PatternLock evaluates imitation and reasoning
over a temporally dense action sequence. For each task, we evaluate 10 episodes
spanning different difficulty levels and repeat each episode three times, for a
total of 30 evaluations per task. We compare against the MemER and
FrameSamp-Modul baselines.
Appendix~\ref{app:real-world-details} provides
further details on the tasks, evaluation protocol, and rollout visualizations.

\begin{figure*}[t]
    \centering
    \includegraphics[width=\textwidth]{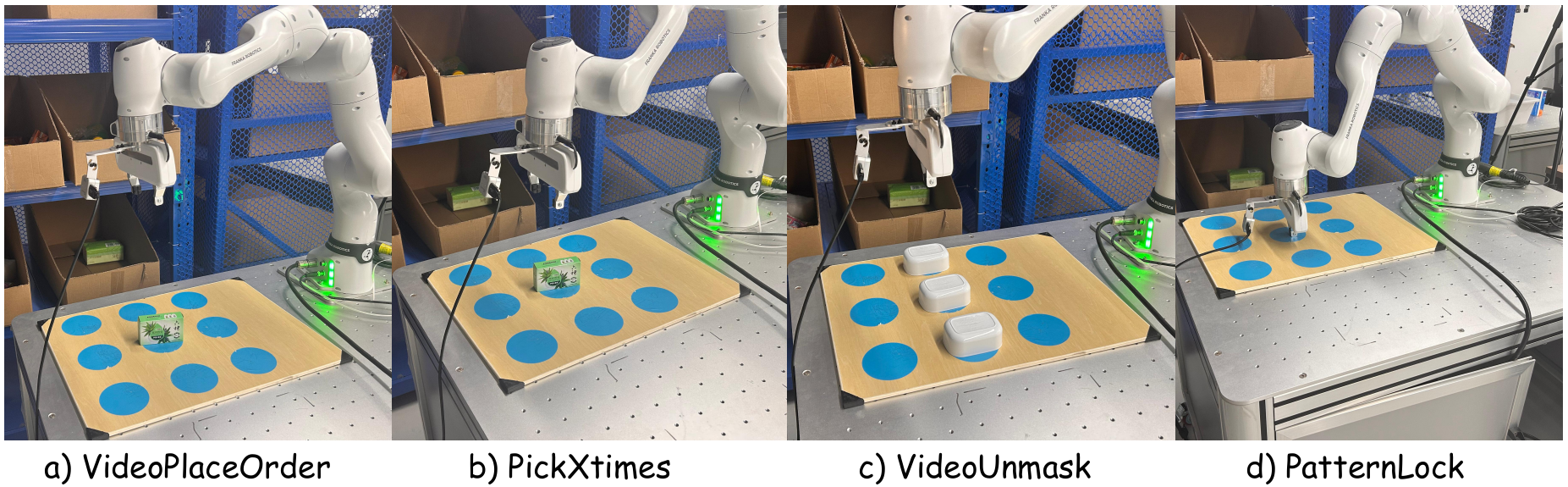}
    \caption{Real-world manipulation tasks used to evaluate \pal{}. The tasks
    require retaining visual evidence from previous observations to support
    subsequent manipulation.}
    \label{fig:real-world-tasks}
\end{figure*}

\begin{table}[t]
\centering
\caption{Real-world success rates (\%, mean $\pm$ sample standard deviation over
three repeats of each 10-episode evaluation).}
\label{tab:real-world-main-results}
\small
\setlength{\tabcolsep}{8pt}
\renewcommand{\arraystretch}{1.12}
\resizebox{\linewidth}{!}{%
\begin{tabular}{lccccc}
\hline
\textbf{Method} & \textbf{PatternLock} & \textbf{PickXtimes} &
\textbf{VideoPlaceOrder} & \textbf{VideoUnmask} & Overall \\
\hline
FrameSamp-Modul & 30.00 $\pm$ 10.00 & 40.00 $\pm$ 30.00 &
13.33 $\pm$ 5.77 & 30.00 $\pm$ 0.00 & 28.33 $\pm$ 3.82 \\
MemER & 33.33 $\pm$ 5.77 & 0.00 $\pm$ 0.00 & 46.67 $\pm$ 30.55 &
\textbf{100.00} $\pm$ \textbf{0.00} & 45.00 $\pm$ 6.61 \\
\hline
\textbf{Ours} & \textbf{40.00} $\pm$ \textbf{0.00} &
\textbf{53.33} $\pm$ \textbf{15.28} & \textbf{50.00} $\pm$ \textbf{0.00} &
\textbf{100.00} $\pm$ \textbf{0.00} & \textbf{60.83} $\pm$ \textbf{3.82} \\
\hline
\end{tabular}%
}
\end{table}
\noindent\textbf{Q4: How does \pal{} perform in real-world robotic manipulation tasks compared with baselines?}
The results are reported in Table~\ref{tab:real-world-main-results}.
FrameSamp-Modul remains
weak across all four tasks and almost entirely fails on VideoPlaceOrder
($13.33\%$), indicating that a short, evenly sampled visual window is insufficient
for recalling occluded events in long videos. MemER performs well on
VideoUnmask ($100.00\%$), a comparatively short-context recall task, but is weaker
than our method on longer-context tasks such as PickXtimes ($0.00\%$ vs.\ $53.33\%$)
and VideoPlaceOrder ($46.67\%$ vs.\ $50.00\%$), and remains limited on PatternLock
($33.33\%$). In contrast, our method attains the highest overall success rate of
$60.83\%$, with the largest gains on the longer-horizon tasks. These results
suggest that converting streaming observations into hierarchical, executable
memory preserves the task-relevant events needed for real-world long-horizon
manipulation, rather than relying on lossy frame sampling or retrieval from raw
history.

\section{Conclusion}

We presented \pal{}, a hierarchical memory pipeline that incrementally converts
streaming visual observations into semantic events and a subtask list
before they are needed for control. By separating early subtask construction
from execution-time selection, \pal{} reduces the need to repeatedly revisit a
large visual history. We further introduced a two-stage training procedure that
combines dense update supervision with verified self-labeling to learn reliable
memory updates under the model's rollout distribution. Experiments on RoboMME
and four real-world manipulation tasks demonstrate that \pal{} improves
memory-dependent robot control, particularly when relevant evidence must be
retained over long horizons.

While \pal{} improves upon prior work, it still has limitations. Its executable
memory is represented as a sequential subtask list and therefore cannot encode
conditional branches during execution. Programmatic planning methods such as
Code as Policies demonstrate that executable programs can express conditional
control flow and perception-dependent behavior \citep{liang2023codepolicieslanguagemodel}.
Future work could extend \pal{} with branchable executable memory that selects
different subtask sequences according to execution-time observations.
\subsection*{AI use statement}

In this work, we used generative AI tools to generate synthetic data sets, design
or provide feedback on research methodology or experiments, implement methods,
assist with translation, clean and reformat datasets, support qualitative and
thematic data analysis, and interpret results. We have not used generative AI
tools to help develop theoretical models or conceptual pipelines, formulate
mathematical claims, or propose or refine hypotheses. Providing critical
ingredients for proving mathematical claims and assisting in the writing of
proofs are not applicable to this work. Additionally, we used generative AI tools to create
or modify scientific figures or images, draft parts of the research paper,
summarize or analyse existing literature, brainstorm, source and search for
information, edit the research paper to improve readability, identify relevant
literature, format references, suggest a structure for the research paper, and
propose a title or keywords for the research paper. We have reviewed all
AI-assisted work: generated data and code were checked by the authors, and all
drafted text was revised by the authors. We take responsibility for the final
content of this work, including text, claims or artifacts produced with the aid
of generative AI.

\subsection*{Ethics statement}
This work does not raise ethical concerns. It studies memory-dependent robot
manipulation in simulation and on a robot arm in a controlled laboratory
setting, involves no human subjects, and we do not foresee harmful applications.

\subsection*{Reproducibility statement}
All code and data needed to reproduce this work are available at
\url{https://qinghangliu.github.io/pal-webpage/}. The streaming memory
update and the Stage~2 self-labeling procedure are given as pseudocode in
Appendix~\ref{alg:streaming-memory-update} and
Appendix~\ref{alg:stage-two-self-labeling}. The network architecture and training
hyperparameters are listed in Appendix~\ref{app:pal-training}. The construction
of the simulated and real-world datasets is described in
Appendices~\ref{app:dataset-construction} and~\ref{app:real-data-cleaning}, and
the evaluation protocols are described in Appendices~\ref{app:simulation-setup}
and~\ref{app:real-world-evaluation}.

\bibliography{iclr2027_conference}
\bibliographystyle{iclr2027_conference}

\appendix
\section{Algorithms}
{\raggedright
This appendix provides the pseudocode for the two procedures described in the
main text. Algorithm~\ref{alg:streaming-memory-update} summarizes one VLM query
and the corresponding updates to the visual, event, and executable memories.
Algorithm~\ref{alg:stage-two-self-labeling} details the self-labeling rollout and
task-grounded verification used in Stage 2.\par}
\medskip
\begin{algorithm}[H]
\caption{Streaming memory update at query $t$}
\label{alg:streaming-memory-update}
\Require{Previous memory $\mathcal{M}_{t-\Delta t}$, visual buffer
         $\mathcal{V}_{t-\Delta t}$, image chunk $\mathcal{C}_t$, instruction $L$}
$\mathcal{V}_t \gets \operatorname{Append}(\mathcal{V}_{t-\Delta t},\mathcal{C}_t)$\;
$(\Delta\mathcal{E}_t,\Delta\mathcal{P}_t,\ell_t)
\gets g_{\theta}(\mathcal{M}_{t-\Delta t},\mathcal{C}_t,L)$\;
$\mathcal{P}_t \gets
\operatorname{Append}(\mathcal{P}_{t-\Delta t},\Delta\mathcal{P}_t)$\;
$\widetilde{\mathcal{E}}_{t-\Delta t} \gets
\begin{cases}
\emptyset, & \Delta\mathcal{P}_t\neq\emptyset,\\
\mathcal{E}_{t-\Delta t}, & \text{otherwise}
\end{cases}$\;
$\mathcal{E}_t \gets
\operatorname{Append}(\widetilde{\mathcal{E}}_{t-\Delta t},\Delta\mathcal{E}_t)$\;
\If{$\Delta\mathcal{E}_t\neq\emptyset$ or $\Delta\mathcal{P}_t\neq\emptyset$}{
    $\mathcal{V}_t\gets\emptyset$\;
}
\Return{$\mathcal{M}_t=(\mathcal{V}_t,\mathcal{E}_t,\mathcal{P}_t)$ and $\ell_t$}\;
\end{algorithm}

\begin{algorithm}[t]
\caption{Stage 2 self-labeling with task-grounded verification}
\label{alg:stage-two-self-labeling}
\Require{Stage 1 model $g_{\theta_1}$, query interval $\Delta t$,
         maximum timestep $T$, training trajectories $\mathcal{T}$ with
         reference updates
         $\{(t_i,\Delta\mathcal{E}^{\mathrm{ref}}_i,\Delta\mathcal{P}^{\mathrm{ref}}_i)\}_{i=1}^{N}$}
\Ensure{Verified self-labeled trajectories $\mathcal{D}_{\mathrm{self}}$}
$\mathcal{D}_{\mathrm{self}}\gets\emptyset$\;
\ForEach{$\tau\in\mathcal{T}$}{
    Initialize $\mathcal{M}_0$ and temporary annotations $\mathcal{D}_{\tau}\gets\emptyset$\;
    $\textit{valid}\gets\textbf{True}$; $t\gets 0$\;
    \While{$t < T$ \textbf{and} $\textit{valid}$}{
        $t\gets t+\Delta t$\;
        $(\Delta\mathcal{E}_t,\Delta\mathcal{P}_t,\ell_t)
        \gets g_{\theta_1}(\mathcal{M}_{t-\Delta t},\mathcal{C}_t,L)$\;
        Find $i$ such that $t\in[\frac{t_{i-1}+t_i}{2},\frac{t_i+t_{i+1}}{2})$\;
        \If{$\Delta\mathcal{E}_t\neq\Delta\mathcal{E}^{\mathrm{ref}}_i$
            \textbf{or} $\Delta\mathcal{E}_t\in\mathcal{E}_{t-\Delta t}$}{
            $\Delta\mathcal{E}_t\gets\emptyset$\;
        }
        \If{$\Delta\mathcal{P}_t\neq\Delta\mathcal{P}^{\mathrm{ref}}_i$
            \textbf{or} $\Delta\mathcal{P}_t\in\mathcal{P}_{t-\Delta t}$}{
            $\Delta\mathcal{P}_t\gets\emptyset$\;
        }
        Update $\mathcal{M}_t=(\mathcal{V}_t,\mathcal{E}_t,\mathcal{P}_t)$ with
        $(\Delta\mathcal{E}_t,\Delta\mathcal{P}_t)$ as in Algorithm~\ref{alg:streaming-memory-update}\;
        Add $(\mathcal{M}_{t-\Delta t},\mathcal{C}_t,L,
        \Delta\mathcal{E}_t,\Delta\mathcal{P}_t,\ell_t)$ to $\mathcal{D}_{\tau}$\;
        \ForEach{unchecked $i$ with $t\geq\frac{t_i+t_{i+1}}{2}$}{
            \If{$\Delta\mathcal{E}^{\mathrm{ref}}_i\notin\mathcal{E}_t$
                \textbf{or} $\Delta\mathcal{P}^{\mathrm{ref}}_i\notin\mathcal{P}_t$}{
                $\textit{valid}\gets\textbf{False}$; \textbf{break}\;
            }
            Mark $i$ as checked\;
        }
    }
    \If{$\textit{valid}$}{
        $\mathcal{D}_{\mathrm{self}}\gets\mathcal{D}_{\mathrm{self}}\cup\{\mathcal{D}_{\tau}\}$\;
    }
}
\Return{$\mathcal{D}_{\mathrm{self}}$}\;
\end{algorithm}

\section{\pal{} Training Details}
\label{app:pal-training}
\subsection{Stage 1 Training Setup}
We build \pal{} upon MemER \citep{sridhar2025memerscalingmemoryrobot}.
Table~\ref{tab:pal-training} summarizes the network architecture and the
Stage 1 training hyperparameters.

\begin{table}[t]
\centering
\caption{Network architecture and Stage 1 training hyperparameters of \pal{}.}
\label{tab:pal-training}
\small
\begin{tabular}{ll}
\toprule
\textbf{Configurations} & \textbf{Values} \\
\midrule
\multicolumn{2}{l}{\textit{Network Architecture}} \\
Base VLM & Qwen3-VL-4B-Instruct \\
Low-level policy & Pretrained MemER VLA \\
Action chunk length & $20$ \\
Fine-tuning & LoRA on language-model linear layers \\
Frozen modules & Vision tower and aligner \\
LoRA rank / $\alpha$ / dropout & $16$ / $32$ / $0.05$ \\
Trainable parameters & $33.0$M ($0.74\%$) \\
\midrule
\multicolumn{2}{l}{\textit{Training Hyperparameters}} \\
Hardware & $8\times$ NVIDIA A800 \\
Parallelism & DeepSpeed ZeRO-2 \\
Optimizer & Fused AdamW \\
Epochs & $8$ \\
Training time & ${\sim}8.9$ hours \\
Learning rate & $8\times10^{-5}$ \\
Schedule & Cosine, warmup ratio $0.05$ \\
Adam $(\beta_1,\beta_2)$ & $(0.9,\ 0.95)$ \\
Weight decay & $0.1$ \\
Gradient clipping & $1.0$ \\
Per-device batch size & $4$ \\
Gradient accumulation & $1$ \\
Global batch size & $32$ \\
Precision & bfloat16 \\
Attention & FlashAttention \\
Max sequence length & $3200$ \\
\bottomrule
\end{tabular}
\end{table}

\subsection{Stage 2 Self-Labeling}
Stage 2 starts from the Stage 1 LoRA adapter and keeps the optimizer,
batch size, and LoRA configuration above. It has two modes, selected according
to how reliable the Stage 1 model already is.

\paragraph{Verifier.}
The Stage 2 verifier checks event and subtask-list updates independently
against their reference updates. All non-coordinate text must match exactly
(100\% correct; paraphrases, omissions, and extra text are rejected). When
coordinates are present, the prediction must include the corresponding
coordinate fields, and each predicted-reference pair must have an L2 distance
of strictly less than
$10$ pixels, i.e., $\|\mathbf{x}^{\mathrm{pred}}-\mathbf{x}^{\mathrm{ref}}\|_2<10$;
without coordinates, the exact-text criterion alone applies.

\begin{figure}[t]
\centering
\includegraphics[width=0.72\linewidth]{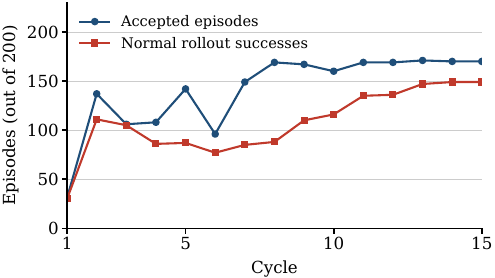}
\caption{Accepted episodes and normal rollout successes over the $15$
self-labeling cycles. Each cycle relabels all $200$ episodes of
PatternLock and VideoPlaceOrder.}
\label{fig:stage2-cycle-acceptance}
\end{figure}

We use one-pass self-labeling for the simulation experiments because the Stage 1
model already produces reliable rollouts: the first pass accepts most training
episodes, so no further labeling rounds are needed. We use cyclic self-labeling
for the real-world experiments because the first pass is less reliable; it
retains only $31$ of $200$ episodes in the first round and improves to $170$ in
the final round. The change in acceptance over cycles is shown in
Figure~\ref{fig:stage2-cycle-acceptance}.

\paragraph{One-pass labeling.}
When the Stage 1 model is already strong enough, we label the training set once
and then train. The Stage 1 model rolls out every training episode, and only
episodes accepted by the verifier in Algorithm~\ref{alg:stage-two-self-labeling}
are kept. The resulting corpus is the accepted self-labeled episodes together
with the original supervised labels of the tasks that are not self-labeled
(BinFill, ButtonUnmask, PickXtimes, StopCube, and SwingXtimes). Non-empty
accepted updates are repeated three extra times, and PatternLock and RouteStick
are upweighted by a factor of four. We then train for $2$ epochs on $8$ GPUs.
In the reported run this corpus contains $178{,}855$ rows and the training run
takes about $5.5$ hours. Of the $100$ training episodes per self-labeled task,
the verifier accepts $100$ for MoveCube, $99$ for PickHighlight and VideoUnmask,
$98$ for InsertPeg, PatternLock, and RouteStick, $92$ for VideoPlaceButton,
$90$ for ButtonUnmaskSwap, $86$ for VideoPlaceOrder and VideoUnmaskSwap, and
$84$ for VideoRepick. The first self-labeling pass already accepts most
episodes of every task, so this single pass provides enough data for training
and no further labeling rounds are needed.

\paragraph{Cyclic labeling.}
When the Stage 1 model is still weak, a single labeling pass accepts too few
episodes, so we alternate training and relabeling. The reported run self-labels
PatternLock and VideoPlaceOrder, both on their conditioning videos, and keeps the
original supervised labels for PickXtimes and VideoUnmask. Both the event and the
subtask list are verified. An initial labeling pass is run with the Stage 1
adapter, after which training proceeds for $3000$ steps in blocks of $200$.
After each block, the current checkpoint relabels every episode of the two
self-labeled tasks; accepted episodes replace their previous labels, while
episodes that fail verification fall back to their original supervised labels.
The next block resumes from that checkpoint and trains on the refreshed corpus.
This run uses $15$ such rounds. An \emph{accepted} episode is one that passes
the verifier of Sec.~4.3 and can therefore be used for training. A
\emph{normal rollout success} is an episode whose final list is already correct
without the verifier. Acceptance rises quickly: the first round keeps
$31$ of $200$ episodes, and the last round keeps $170$. Non-empty accepted
updates are repeated three extra times, as in the one-pass mode. The final
corpus contains $40{,}144$ rows, and the $15$ training rounds take about
$0.7$ hours in total.
Figure~\ref{fig:stage2-cycle-acceptance} shows how both quantities change over
the $15$ cycles.

\begin{table*}[t]
\centering
\caption{Full RoboMME results. Success rates (\%) are reported for all 16
tasks, grouped by how the executable subtask list is obtained. The highest
score for each task is in bold.}
\label{tab:simulation-full-results}
\setlength{\tabcolsep}{1.8pt}
\renewcommand{\arraystretch}{1.08}
\footnotesize
\resizebox{\textwidth}{!}{%
\begin{tabular}{l|rrrrrrr|rr|rrrr|rcr|r}
\toprule
\multirow{2}{*}{\textbf{Method}}
& \multicolumn{7}{c}{\textbf{Short-video}}
& \multicolumn{2}{c}{\textbf{Long-video}}
& \multicolumn{4}{c}{\textbf{In-context}}
& \multicolumn{3}{c}{\textbf{Episode-start}}
& \multirow{2}{*}{\textbf{AVG}} \\
\cmidrule(lr){2-8}\cmidrule(lr){9-10}\cmidrule(lr){11-14}\cmidrule(lr){15-17}
& \shortstack{Video\\Umsk} & \shortstack{Video\\Repick} & \shortstack{Video\\UmskS}
& \shortstack{Pattern\\Lock} & \shortstack{Route\\Stick} & \shortstack{Move\\Cube}
& \shortstack{Insert\\Peg} & \shortstack{Video\\PlcBtn} & \shortstack{Video\\PlcOrd}
& \shortstack{Button\\Umsk} & \shortstack{Button\\UmskS} & \shortstack{Pick\\HighL}
& \shortstack{Stop\\Cube} & \shortstack{Pick\\Xtimes} & \shortstack{Bin\\Fill}
& \shortstack{Swing\\Xtimes} & \\
\midrule
\multicolumn{18}{l}{\textbf{\textit{Baselines}}} \\
$\pi_{0.5}$ & 20.44 & 0.44 & 18.67 & 2.89 & 4.67 & 26.00 & 1.56 & 31.11 & 25.78 & 22.22 & 6.67 & 11.33 & 6.67 & 42.89 & 30.00 & 35.56 & 17.93 \\
EventVLA & 66.00 & 30.00 & 17.60 & 4.00 & 27.60 & 44.00 & 4.00 & 22.00 & 16.00 & 60.00 & 12.00 & 12.00 & \textbf{46.00} & 48.00 & 6.00 & 10.00 & 26.57 \\
FrameSamp-Modul & 32.67 & 30.44 & 24.44 & \textbf{53.56} & \textbf{66.67} & 77.78 & 7.56 & 60.00 & 32.00 & 25.11 & 18.22 & 22.89 & 42.00 & 87.33 & 39.56 & \textbf{92.00} & 44.51 \\
MemER & 81.33 & 25.33 & \textbf{38.00} & 16.67 & 12.00 & \textbf{82.67} & 6.67 & 30.00 & 26.00 & 72.00 & \textbf{21.33} & \textbf{70.67} & 0.00 & 79.33 & \textbf{56.67} & 59.33 & 42.38 \\
GPT-6-astra + VLA & 78.00 & \textbf{32.00} & 28.00 & 24.00 & 4.00 & 54.00 & 2.00 & 50.00 & 82.00 & 32.00 & 0.00 & \textbf{74.00} & 6.00 & \textbf{88.00} & 38.00 & 44.00 & 39.75 \\
\midrule
\multicolumn{18}{l}{\textbf{\textit{Streaming Reasoning}}} \\
Stage 1 Only & 70.80 & 30.80 & 17.60 & 33.60 & 22.00 & 42.40 & 2.80 & 18.80 & 16.00 & 62.40 & 12.38 & 48.40 & 6.80 & 86.40 & 51.61 & 74.80 & 37.35 \\
Sparse Label & 76.40 & 28.40 & 18.80 & 24.40 & 16.80 & 81.20 & \textbf{11.20} & 40.88 & 20.00 & 74.91 & 19.24 & 58.00 & 2.40 & 87.60 & 56.40 & 77.60 & 43.39 \\
\textbf{\pal{} (ours)} & \textbf{95.20} & \textbf{32.00} & 35.20 & 50.80 & 22.40 & 76.40 & 4.80 & \textbf{80.00} & \textbf{83.20} & \textbf{90.80} & 20.40 & 68.40 & 5.20 & 87.20 & 65.20 & 79.60 & \textbf{56.05} \\
\bottomrule
\end{tabular}%
}
\end{table*}

\section{Simulated Experimental Details}
\subsection{Full Simulation results}
\label{app:simulation-results}
The per-task results in Table~\ref{tab:simulation-full-results} show that \pal{}
performs best on VideoUnmask ($95.20\%$), ButtonUnmask ($90.80\%$),
VideoPlaceButton ($80.00\%$), and VideoPlaceOrder ($83.20\%$). The margins
over the strongest baseline are $13.87$, $18.80$, $20.00$, and $39.20$
percentage points, respectively.
\pal{} is comparable to the strongest baseline on PatternLock ($50.80\%$ vs.\
$53.56\%$), MoveCube ($76.40\%$ vs.\ $82.67\%$), PickHighlight
($68.40\%$ vs.\ $70.67\%$), and PickXtimes ($87.20\%$ vs.\ $87.33\%$).
It also leads BinFill with $65.20\%$, compared with $56.67\%$ for MemER.

\pal{} remains weaker on RouteStick ($22.40\%$ vs.\ $66.67\%$) and
SwingXtimes ($79.60\%$ vs.\ $92.00\%$). Its largest limitations are
InsertPeg ($4.80\%$) and StopCube ($5.20\%$), where the best alternative
scores are $11.20\%$ and $46.00\%$, respectively. VideoRepick and the two
swap tasks also remain difficult: the best scores are $32.00\%$ on
VideoRepick, $38.00\%$ on VideoUnmaskSwap, and $21.33\%$ on
ButtonUnmaskSwap.

\subsection{Dataset Construction}
\label{app:dataset-construction}
We construct sparse labels from each simulator episode. A sparse label is
created only when the event memory or the executable \texttt{subtask\_list}
should change. The way we construct the list depends on the task type. We use
the simulator's oracle event annotations only to determine what happened in the
episode and when the corresponding information becomes available; the model is
still trained to infer the event from visual observations.

We use the following three cases:
\begin{enumerate}
    \item \textbf{Episode-start tasks.} The ordered
    \texttt{subtask\_list} can be determined at the beginning of the episode.
    We therefore construct the full list at the first query and keep it
    unchanged during the rest of the episode. These tasks do not need an
    intermediate event label for constructing the list.

    \item \textbf{Tasks with changing environments.} We first read the oracle
    event sequence from the simulator. For each task, a task-specific
    rule-based procedure determines when an event can be inferred from the
    observations and which subtask or subtasks should be added to the
    \texttt{subtask\_list}. Before enough evidence is available, the list is
    left unchanged. After the event can be inferred, the sparse label first
    records the event; when the information is sufficient to specify the next
    action, a later sparse label constructs or extends the
    \texttt{subtask\_list}. The rules for this timing and mapping are defined
    separately for each task.

    \item \textbf{StopCube.} This task does not require a separate event
    annotation. At each timestep, we set the executable
    \texttt{subtask\_list} to the current subtask itself, so the list and the
    subtask are identical.
\end{enumerate}

Thus, depending on the task, a sparse sample can supervise an event update, a
\texttt{subtask\_list} update, or both. Between two update points, the memory
state is kept unchanged. Stage 1 subsequently expands these sparse update
points into a timing band with multiple query timesteps, as described in
Sec.~\ref{methodology}; this expansion is separate from the sparse-label
construction above.

\subsection{Implementation Details and Baselines}
\label{app:simulation-setup}
We first describe the \pal{} implementation and evaluation protocol, followed by
the configurations of the comparison methods.

\paragraph{\pal{}.}
The \pal{} training setup is described in Appendix~\ref{app:pal-training}.
Table~\ref{tab:pal-inference} summarizes the inference and evaluation protocol.
We evaluate \pal{} with 50 episodes per task for each seed. For tasks with a
conditioning video, we feed the video to the robot as a stream, even though the
entire video is available at the beginning. This setting is practically
meaningful because it evaluates the model's ability to reason in real time.

\begin{table}[t]
\centering
\caption{\pal{} inference and evaluation protocol on RoboMME.}
\label{tab:pal-inference}
\small
\begin{tabular}{ll}
\toprule
\textbf{Configurations} & \textbf{Values} \\
\midrule
Conditioning-video sampling & Every $4$ timesteps \\
Execution action chunk & $16$ \\
VLM query & Once per action chunk \\
Visual buffer input & $4$ uniformly sampled frames \\
Policy seeds & $42, 43, 44, 45, 46$ \\
\bottomrule
\end{tabular}
\end{table}

\paragraph{Baselines.}
For the $\pi_{0.5}$, MemER, and FrameSamp-Modul baselines, we use the results
reported by RoboMME~\citep{dai2026robommebenchmarkingunderstandingmemory};
the original paper evaluates each method using three policy checkpoints and
three seeds per checkpoint, yielding nine evaluation runs in total.

\paragraph{EventVLA.}
We adapt the original EventVLA implementation by using the frames annotated as
event or \texttt{subtask\_list} updates in \pal{} as keyframes. We also implement a
streaming keyframe-selection procedure that selects keyframes from the history
video as it is observed. All other settings follow the default EventVLA
configuration. For long-horizon tasks, we increase the maximum number of stored
keyframes to eight to accommodate tasks that require more historical evidence.
The model is trained for 100k steps, with a maximum of eight keyframes at
inference time.

\paragraph{GPT-6-astra + VLA.}
We adopt a structure similar to that of \citep{torne2026memmultiscaleembodiedmemory},
in which memory is maintained as a language description of the current
environment. At each query, the VLM reads the four most recent frames and decides
the next subtask for the low-level VLA to execute. For the conditioning video, we
sample one frame every 33 frames, using at most 24 frames; for videos shorter
than 100 frames, we provide at least three frames. This baseline uses the same
low-level VLA as \pal{} but is evaluated with one seed because repeated API calls
incur substantial evaluation time.

\subsection{More Analysis}
\subsubsection{GPT-6-astra + VLA}
We analyze the results of GPT-6-astra + VLA to identify its failure modes.

\textbf{Failure mode 1: insensitivity to subtle environmental changes.}
The model cannot reliably detect small changes in the scene. For example, whether
a button has been pressed is visually subtle, so the model often cannot judge
whether the press succeeded and therefore does not output the next subtask. This
is reflected in its failures on ButtonUnmask and ButtonUnmaskSwap, where it
performs notably worse than on the similar tasks VideoUnmask and
VideoUnmaskSwap, which do not require pressing a button.

\textbf{Failure mode 2: loss of rapidly changing historical information.}
GPT-6-astra achieves the second-best performance on the Long-video tasks, yet
fails on Short-video tasks such as RouteStick and PatternLock. These tasks are
information-dense and their evidence spans only a short interval, so uniform
downsampling does not retain enough information.

\textbf{Failure mode 3: unstable high-level--low-level interface.}
The model does not interface reliably with the low-level policy. On tasks such as
BinFill and VideoUnmask, the high-level VLM must specify which cube or box to
pick. However, the bounding box it outputs jumps frequently from one location to
another, which interrupts the motion of the low-level VLA.

\subsubsection{\pal{}}
\pal{} also performs poorly on InsertPeg, StopCube, and RouteStick. InsertPeg
requires precise manipulation to align and insert the peg. In the $250$
evaluation episodes, $101$ ended in timeout because the peg was not inserted
within the allowed horizon, contributing to the low success rate of $4.8\%$.
StopCube requires the robot to anticipate the motion of a dynamically moving
object and react quickly enough to press the button at the right moment. These
fine-control and rapid-reaction requirements are outside the main scope of this
work, which focuses on memory update and long-horizon reasoning. RouteStick
exposes a different limitation: the high-level subtask selection and the
low-level action execution are not always well coordinated. In an auxiliary
experiment, reducing the action-chunk size from $16$ to $8$ increased the
RouteStick success rate from $22.4\%$ to $46.0\%$, suggesting that finer
execution granularity substantially improves this coordination.
\section{Real-world Experimental Details}
\label{app:real-world-details}

\subsection{Data Collection and processing}
\label{app:real-data-cleaning}
We first use gripper actions and intervals in which the robot arm remains
stationary to obtain coarse temporal annotations of each trajectory, providing
initial estimates of action and subtask boundaries without frame-level manual
annotation. We then provide these coarse annotations together with the
recorded trajectory videos to GPT-5.6-Sol for fine-grained annotation; GPT-5.6-Sol
refines the temporal boundaries and produces the task-level labels used to construct the
event and executable subtask supervision. Finally, we remove the stationary
portions identified during this process while preserving frames around gripper
transitions, subtask boundaries, and task completion, synchronize the remaining
front and wrist RGB observations with the robot states and actions, crop and
resize both views using fixed camera-specific regions, and re-index the
trajectory. Each retained action is rewritten to target the next retained
observation, yielding the final action-aligned trajectories used in our
experiments.

\subsection{Settings and Implementation Details}
\label{app:real-world-tasks}

Table~\ref{tab:real-world-task-difficulty} summarizes how task difficulty is
varied in our real-world evaluation. Difficulty is controlled by the amount of
task-relevant history that must be retained. 
In PickXtimes, the robot repeatedly picks up a green box
and returns it to its original position, so the required number of successful
picks must be remembered. In VideoPlaceOrder, the robot watches a video in which
a green box is placed at several locations and then moves the box to the location
specified by its order in the demonstration. In PatternLock, the robot watches a
path over a $3\times3$ grid and then reproduces the same sequence of movements.
In VideoUnmask, the robot watches three colored cubes being covered, remembers
the color-position mapping, and then points to the requested cube; this task has
no difficulty levels. The training data, evaluation trajectories, and task
configurations are generated using GPT-6-astra. We collect 100 training episodes
for each task and 200 episodes for VideoPlaceOrder because each VideoPlaceOrder
episode is relatively short.

We use a Franka Research 3 robot arm equipped with a front-facing camera and a
wrist-mounted camera. Inference runs on a single NVIDIA RTX 4090 GPU, and the
inference computer communicates with the robot through an SSH-based ROS bridge.
The robot is controlled by a low-level impedance controller that executes the
VLA-generated commands at approximately 950~Hz. Excluding network latency, the
VLA generates action commands at approximately 10--12~Hz.

\begin{table}[t]
\centering
\caption{Difficulty settings for the real-world tasks. PickXtimes uses three
levels, while VideoUnmask has no difficulty distinction.}
\label{tab:real-world-task-difficulty}
\small
\setlength{\tabcolsep}{4pt}
\begin{tabular}{>{\raggedright\arraybackslash}p{0.18\linewidth}
>{\centering\arraybackslash}p{0.18\linewidth}
>{\centering\arraybackslash}p{0.18\linewidth}
>{\centering\arraybackslash}p{0.18\linewidth}
>{\centering\arraybackslash}p{0.18\linewidth}}
\hline
\textbf{Task} & \textbf{Level 1} & \textbf{Level 2} & \textbf{Level 3} &
\textbf{Level 4} \\
\hline
PickXtimes & 2 successful picks &
3 successful picks &
4 successful picks &
-- \\
\hline
VideoPlaceOrder & 2 placements &
3 placements &
4 placements &
5 placements \\
\hline
PatternLock & 1 movement &
2 movements &
3 movements &
4 movements \\
\hline
VideoUnmask & \multicolumn{4}{c}{No difficulty levels} \\
\hline
\end{tabular}
\end{table}

\subsection{Evaluation Protocol}
\label{app:real-world-evaluation}
We apply the same task-specific evaluation criteria to all methods to ensure fair
comparisons. VideoPlaceOrder, PatternLock, and VideoUnmask use pre-recorded
conditioning videos to reduce the manual effort required during evaluation.
For PickXtimes, Levels 1--3 contain 5, 3, and 2 episodes, respectively. For
VideoPlaceOrder and PatternLock, Levels 1--4 contain 3, 3, 2, and 2 episodes,
respectively. VideoUnmask has no difficulty levels and uses 10 episodes under
its fixed setting. Each episode is repeated three times, yielding 30 evaluations
per task.
Across all tasks, an episode is marked as a failure if it exceeds the limit of
1,300 timesteps or triggers the robot arm's safety lock.

\paragraph{PickXtimes.}
An episode is successful if the robot lifts the green box off the table exactly
the specified number of times and subsequently keeps its arm raised for 100
timesteps without reaching down to pick up the box again. Any additional contact
with the box after the required number of repetitions is counted as a failure.
To reduce evaluation time, a trial is also terminated and counted as a failure
if any individual pick-and-place cycle takes more than 650 timesteps. The robot
arm and the green box start from the same fixed positions in every episode.

\paragraph{VideoPlaceOrder.}
An episode is successful if the green box is on the designated target area when
the gripper releases it. A placement is considered valid if the box's vertical
projection overlaps the target area. An episode fails if the box is outside all
target areas upon release or touches an incorrect target area. The green box is
initialized at the center of the workspace, and the robot arm starts from the
same fixed pose in every episode.

\paragraph{PatternLock.}
An episode is successful if the gripper visits the grid points in the prescribed
order. A grid point is considered visited when any part of the gripper passes
directly above it. An episode fails if the gripper leaves the grid region or
passes over an incorrect grid point. At the beginning of each episode, the
gripper is closed and the robot arm is placed at the initial point of the target
trajectory.

\paragraph{VideoUnmask.}
An episode is successful if the robot points to the correct box after the
demonstration video. An episode is failed if the robot points to the wrong box.
The robot arm starts from the same fixed pose in every
episode.
\subsection{Rollout visualization}
\label{app:real-world-rollouts}
\begin{figure*}[t]
    \centering
    \textbf{PatternLock}\\[2pt]
    \includegraphics[width=\textwidth]{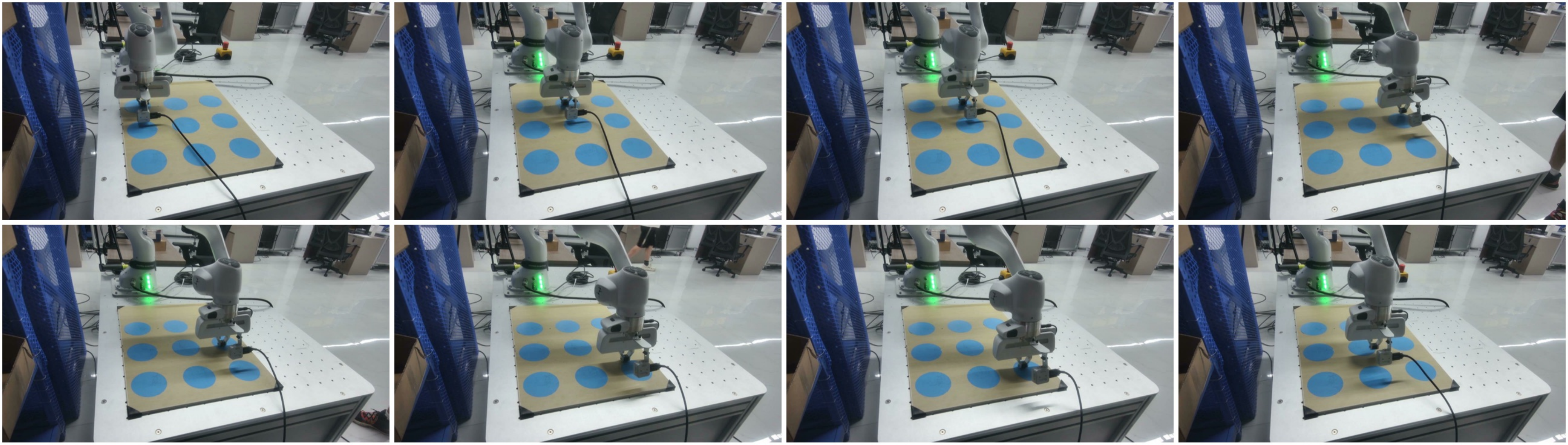}\\[4pt]
    \textbf{PickXtimes}\\[2pt]
    \includegraphics[width=\textwidth]{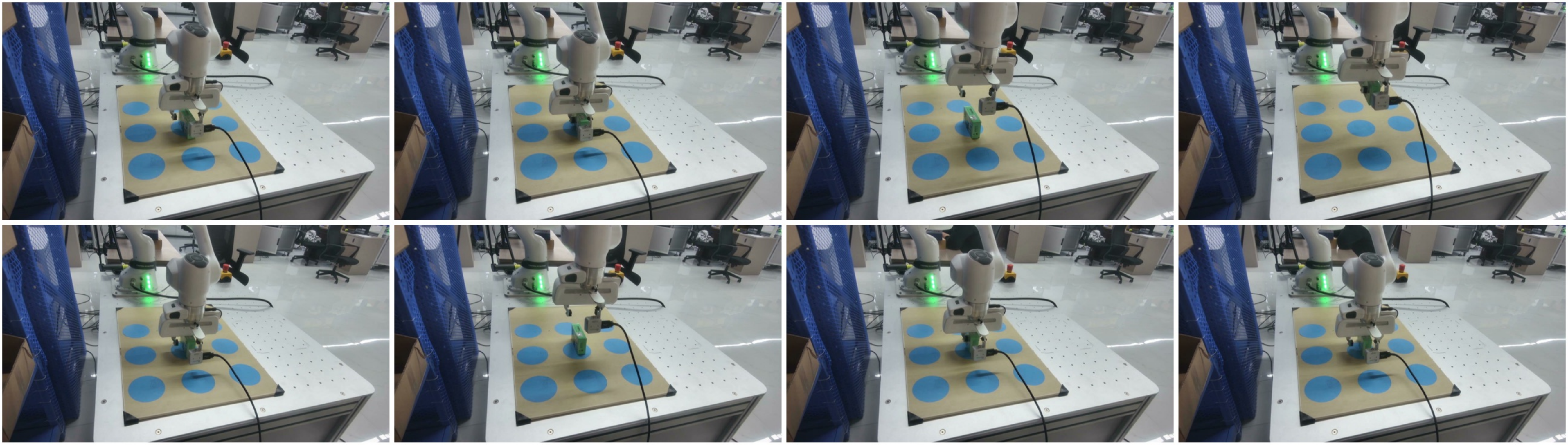}\\[4pt]
    \textbf{VideoPlaceOrder}\\[2pt]
    \includegraphics[width=\textwidth]{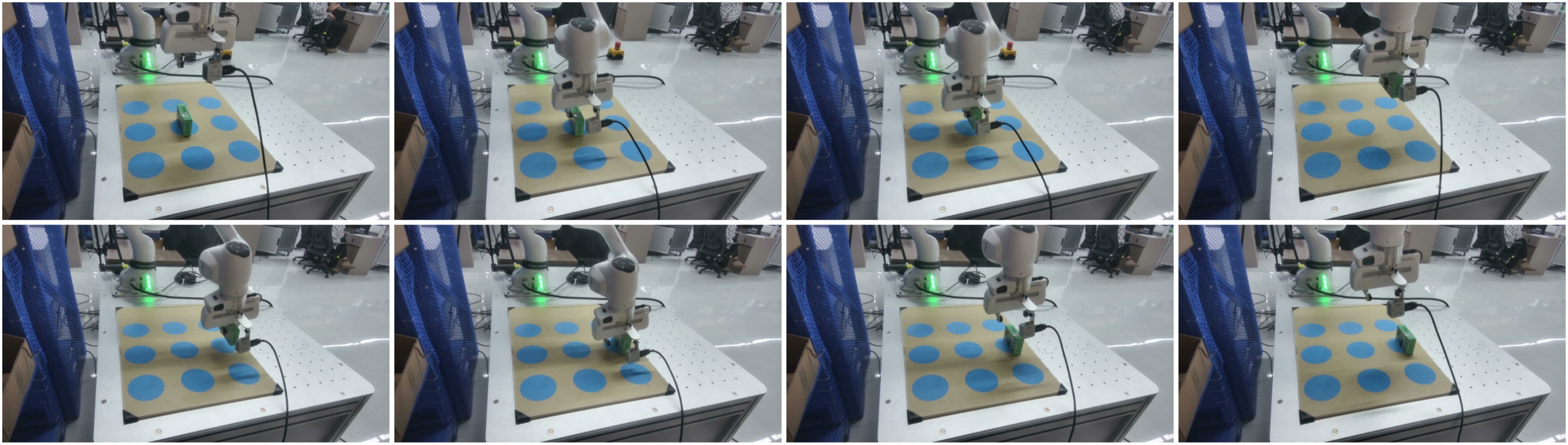}\\[4pt]
    \textbf{VideoUnmask}\\[2pt]
    \includegraphics[width=\textwidth]{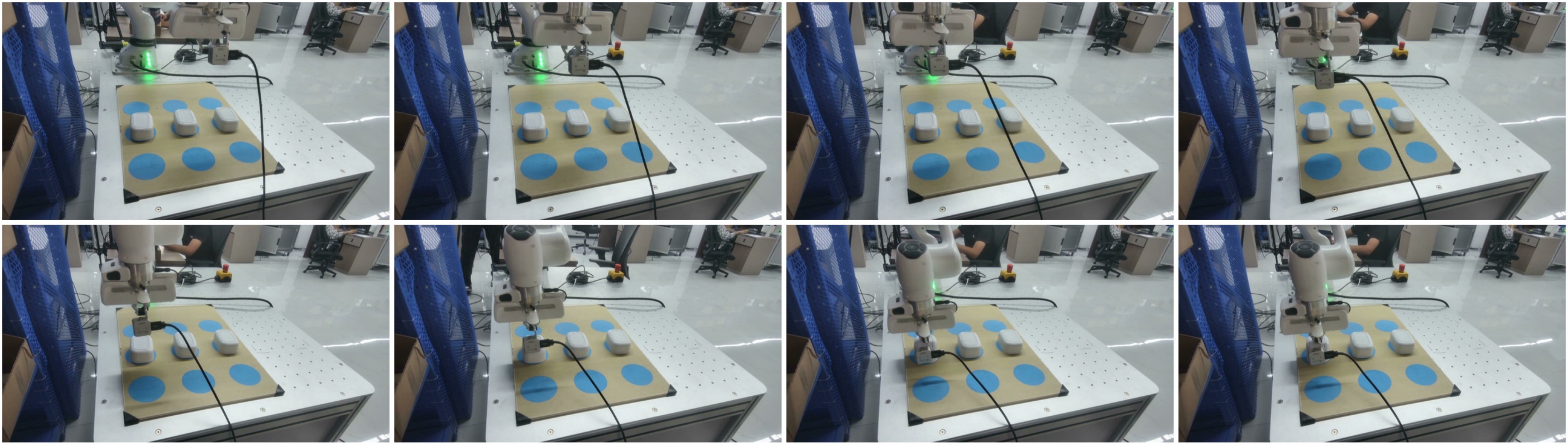}
    \caption{Real-world rollout visualizations of \pal{}. From top to bottom, the
    rows show PatternLock, PickXtimes, VideoPlaceOrder, and VideoUnmask, with
    eight frames sampled along each rollout.}
    \label{fig:real-world-rollouts}
\end{figure*}
Fig.~\ref{fig:real-world-rollouts} shows representative real-world rollouts of
\pal{}, with one task per row and eight frames sampled along each rollout. In
PatternLock, the robot watches a path over the $3\times3$ grid and then
reproduces the same sequence of movements. In PickXtimes, the robot repeatedly
picks up the green box and returns it to its original position, stopping after
the specified number of picks. In VideoPlaceOrder, the robot watches a video in
which a green box is placed at several locations and then moves the box to the
location specified by its order in the demonstration. In VideoUnmask, the robot
watches three colored cubes being covered and then points to the requested cube.

\subsection{More Analysis}
\label{app:real-world-more-analysis}
The baseline failures reveal distinct limitations in task-progress tracking and
visual target grounding (Table~\ref{tab:real-world-main-results}). On
PickXtimes, MemER achieves $0.00\%$ success because it fails to track how many
pick cycles have been completed and repeatedly outputs the initial subtask
instead of advancing to the next one. On VideoUnmask, FrameSamp-Modul achieves
only $30.00\%$ success, close to the $33.33\%$ expected from randomly choosing one
of the three target boxes. In the observed rollouts, it repeatedly points to
the middle box rather than grounding its choice in the requested cube.

\section{Prompt}
\label{app:prompt}
\begin{figure}[t]
\centering
\includegraphics[width=\linewidth]{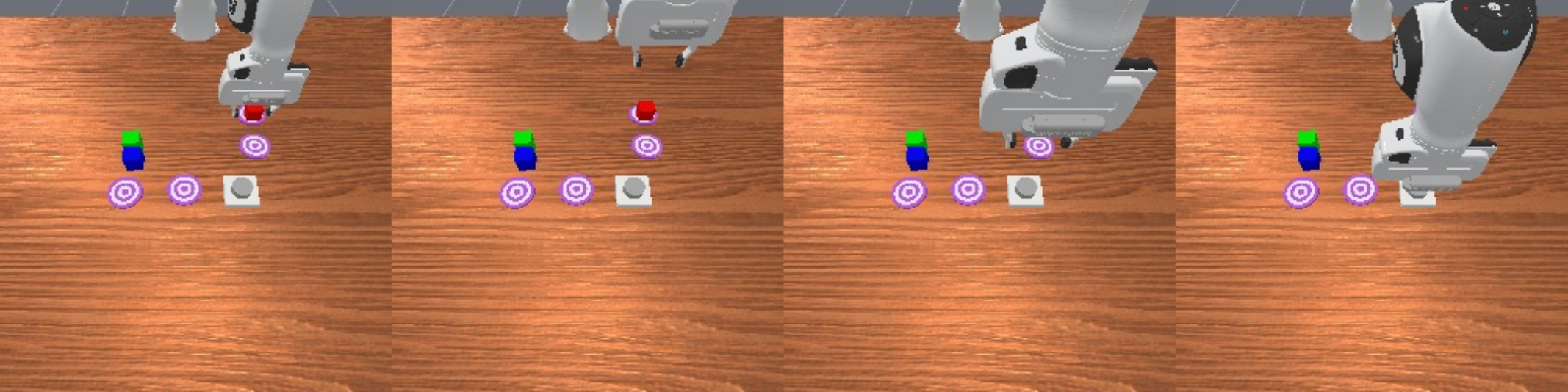}
\caption{The four conditioning images used in the video-phase prompt above, in
timestep order: $349$, $378$, $407$, and $436$.}
\label{fig:prompt-example}
\end{figure}

\begin{figure*}[t]
\centering
\includegraphics[width=\textwidth]{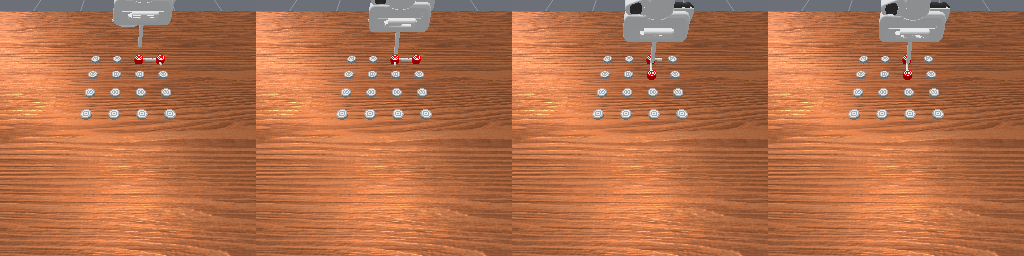}
\caption{The four execution images used in the execution-phase prompt above, in
timestep order: $58$, $71$, $83$, and $96$.}
\label{fig:prompt-example-execution}
\end{figure*}

We show two inference queries. The first is from the video phase of
VideoPlaceOrder, where the model updates the subtask list from the conditioning
video. The second is from the execution phase of PatternLock, where the model
selects the current subtask from the subtask list. In both, the system message
fixes the output schema, the user message gives the task, the current memory, and
the selected frames, and the output is the model's response at this query. The four images of each query are supplied at the timesteps marked below and are
shown in Fig.~\ref{fig:prompt-example} and
Fig.~\ref{fig:prompt-example-execution}; their file paths are omitted.

\lstset{basicstyle=\ttfamily\scriptsize,breaklines=true,breakatwhitespace=true,
columns=fullflexible,frame=single,xleftmargin=0.5em,xrightmargin=0.5em}
\begin{lstlisting}
System:
Update the robot subtask list from selected conditioning images and inferred pre-phase events.
Report the completed event inferred from the images; do not predict the next event.

Return JSON with:
- subtask_list: ordered subtasks inferred so far
- previous_phase_event: newly completed event for the current timestep, or null when there is no new event; do not copy an inferred event

User:
Task goal: watch the video carefully, then place the red cube on the second target it was previously placed on
Phase: video watching
Visual context: front-view camera; left/right are mirrored, so reason in robot-centered coordinates.
Use image coordinates as [y, x] with origin at the top-left.
Previous subtask list, last updated unknown: none
Inferred pre-phase events at sampled memory time, last updated 2.90 seconds (87 timesteps) ago:
1. pick up the cube at <bbox>
2. drop the cube onto target at <bbox>
3. pick up the cube at <bbox>
4. drop the cube onto target at <bbox>
Images are selected around event or subtask-list update times and ordered by timestep.
Question: Update the subtask list from the selected conditioning images.
Grounded object state is included with the selected images.
Selected conditioning images:
Image 1 at timestep 349: <image>
Image 2 at timestep 378: <image>
Image 3 at timestep 407: <image>
Image 4 at timestep 436: <image>

Output:
{"subtask_list": "1. pick up the cube at <ungrounded> 2. drop the cube onto target at <bbox>", "previous_phase_event": "press the button at <bbox>"}
\end{lstlisting}

\begin{lstlisting}
System:
You select the robot's current executable subtask after conditioning is complete.

Decision rule: choose current_subtask from the subtask list using selected execution images. Keep subtask_list unchanged unless the visual evidence clearly shows it is incomplete or wrong. When unchanged, return subtask_list as null instead of copying the previous list.

Return JSON using only the fields supervised for this sample; not every field is required:
- subtask_list (optional): ordered subtasks for the task
- current_subtask: executable action at the current timestep
- previous_phase_event (optional): newly completed task-specific event, or null when unchanged

User:
Task goal: watch the video carefully, then use the stick attached to the robot to retrace the same pattern
Phase: execution
Visual context: front-view camera; left/right are mirrored, so reason in robot-centered coordinates.
Use image coordinates as [y, x] with origin at the top-left.
Subtask list, last updated unknown: 1. move backward 2. move right 3. move forward 4. move backward-right
Previous query current_subtask: 3. move forward
Inferred pre-phase events at sampled memory time, last updated unknown:
none
Selected execution images are sampled from the current visual window through the current timestep.
Question: Predict the current executable subtask.
Grounded object state is included with the selected images.
Selected execution images:
Image 1 at timestep 58: <image>
Image 2 at timestep 71: <image>
Image 3 at timestep 83: <image>
Image 4 at timestep 96: <image>

Output:
{"current_subtask": "4. move backward-right"}
\end{lstlisting}

\end{document}